%% file: root.tex
\documentclass[letterpaper, 10 pt, conference]{ieeeconf}  % Comment this line out if you need a4paper

\IEEEoverridecommandlockouts                              % This command is only needed if 
\usepackage[english]{babel}
\usepackage[utf8]{inputenc}

\usepackage{amsmath}
\usepackage{amssymb}

\usepackage{caption}
\usepackage{subcaption}

\usepackage{booktabs}
\usepackage{url}

\usepackage[colorinlistoftodos]{todonotes}

\usepackage[
    ruled,
    vlined,
    linesnumbered
]{algorithm2e}

\newcommand{\eg}{\textit{e.g.}}

\newcommand{\etal}{\textit{et al.}}

\title{\LARGE \bf
Reward-Guided RRT* for Budgeted Task and Motion Planning
}

\author{Vinícius Araújo de Oliveira$^{1}$, Gabriel Ferreira Mendes
      Lourenço$^{1}$,\\Paulo L. J. Drews-Jr$^{2}$, Armando Alves Neto$^{3}$, and Douglas G. Macharet$^{1}$% <-this % stops a space
\thanks{This work was supported by CAPES/Brazil - Finance Code 001, CNPq/Brazil, and FAPEMIG/Brazil.}% <-this % stops a space
\thanks{$^{1}$V. A. Oliveira, G. F. M. Lourenço, D. G. Macharet are with the Computer Vision and Robotics Laboratory (VeRLab), Dept. of Computer Science, Universidade Federal de Minas Gerais, Belo Horizonte, Brazil. Emails: {\tt\small viniciusoliveira2006@ufmg.br, gabrielferreira10@ufmg.br, doug@dcc.ufmg.br}}
\thanks{$^{2}$P. L. J. Drews-Jr is with the NAUTEC, Center for Computational Science, Universidade Federal do Rio Grande - FURG, Rio Grande, Brazil. Email: {\tt\small paulodrews@furg.br}}%
\thanks{$^{3}$A. A. Neto is with the Dept. of Electronics Engineering, Universidade Federal de Minas Gerais, Belo Horizonte, Brazil. Email: {\tt\small aaneto@cpdee.ufmg.br}}%
}

\begin{document}

\maketitle
\thispagestyle{empty}
\pagestyle{empty}

%%%%%%%%%%%%%%%%%%%%%%%%%%%%%%%%%%%%%%%%%%%%%%%%%%%%%%%%%%%%%%%%%%%%%%%%%%%%%%%%
\begin{abstract}
Robotic missions involving spatially distributed tasks often require jointly selecting rewarding locations and planning feasible motions under limited travel budgets. This paper proposes Reward-Guided RRT*, a new sampling-based framework that constructs a budget-feasible mission through a sequence of locally joint task-selection-and-motion-planning stages, evaluated directly in the continuous state space, without requiring a precomputed task-transition graph. The method incrementally builds a mission using reward-guided RRT* searches, while a return-feasibility procedure ensures that each selected task can be visited without violating the mission budget. Experiments with increasing obstacle complexity and number of tasks show solutions comparable to VNS-PRM* while requiring substantially fewer sampled states. We also extend the framework to Dubins steering and evaluate it in CoppeliaSim with a scaled car-like robotic platform, demonstrating its ability to incorporate vehicle kinematic constraints.
\end{abstract}

\input{defs}

%%%%%%%%%%%%%%%%%%%%%%%%%%%%%%%%%%%%%%%%%%%%%%%%%%%%
\section{Introduction}

Many robotic applications require an autonomous vehicle to perform a collection of spatially distributed tasks while operating under limited resources. Examples include inspection, environmental monitoring, search and rescue, and data collection. In these scenarios, the vehicle must simultaneously determine \emph{where} to perform tasks and \emph{how} to move between them. Visiting additional task locations may increase the mission reward, but it also consumes the travel budget and can make it impossible to reach the final destination. This coupling between discrete task selection and continuous motion planning becomes particularly challenging in environments with obstacles, nonholonomic constraints, or other motion constraints.

\begin{figure}[t]
    \centering
    \includegraphics[width=.65\linewidth]{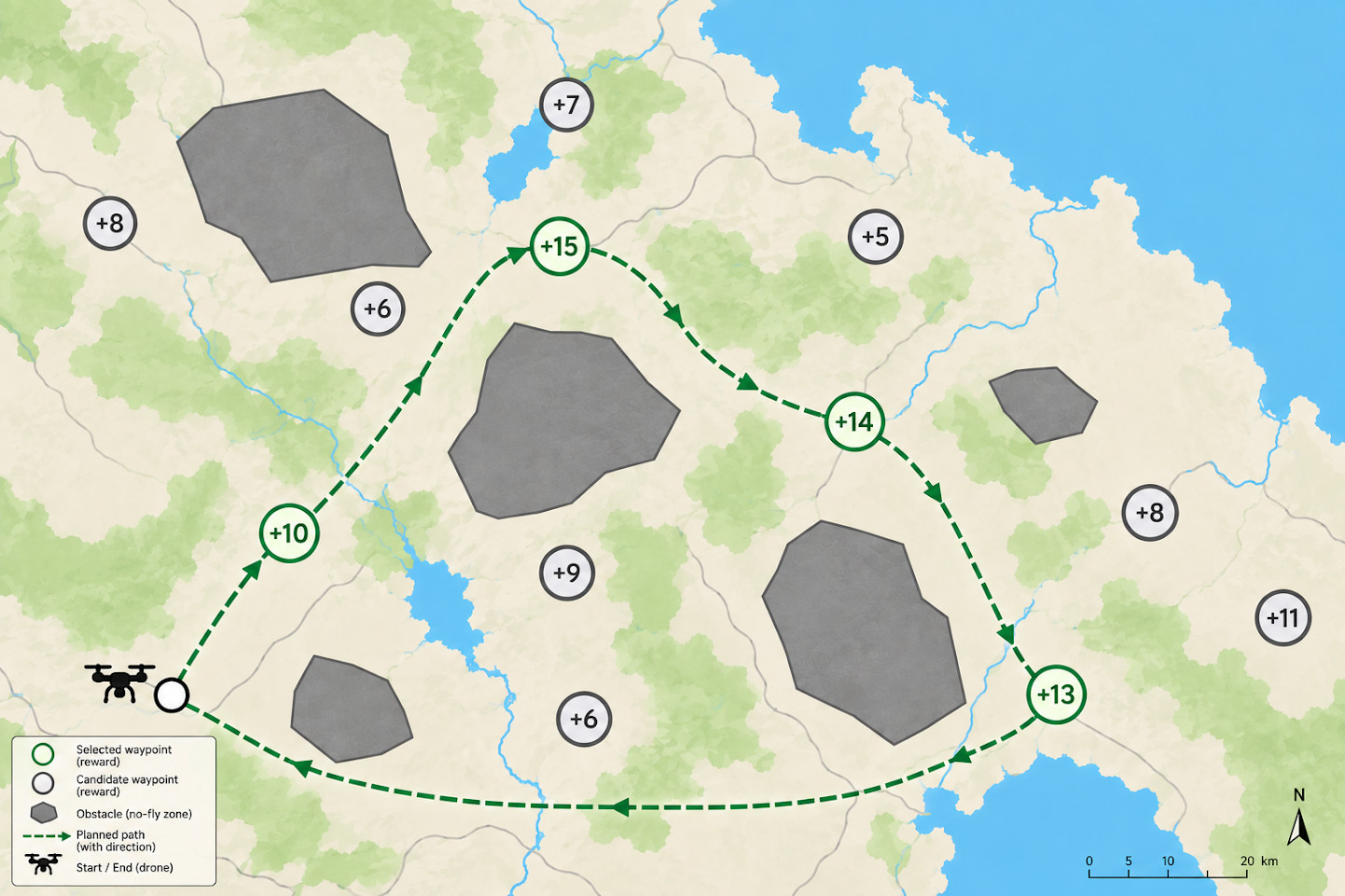}
    \caption{Illustration of budgeted task and motion planning, jointly selecting reward-bearing tasks and collision-free paths to maximize reward under a limited travel budget.}
    \label{fig:introduction}
    \vspace{-5mm}
\end{figure}

In this context, the Orienteering Problem (OP)~\cite{golden1987} provides a natural abstraction for this class of missions. Given a set of reward-bearing locations and a limited travel budget, the objective is to select and visit a subset of locations that maximizes the collected reward (Fig.~\ref{fig:introduction}). Classical OP methods, however, assume that the underlying graph and its travel costs are available beforehand \cite{gunawan2016}. In Robotics, building such a graph may require a large number of pairwise motion-planning queries and may provide only a discrete approximation of the continuous state space. Moreover, task selection and motion planning are typically treated as separate stages, despite their strong interdependence.

This paper investigates an alternative sampling-based scheme in which promising task locations and feasible motions are discovered jointly. Rather than first constructing a complete roadmap and subsequently solving a discrete OP, we incrementally grow a search tree from the initial state while directing exploration toward reward-bearing locations. Each branch represents a feasible partial mission, including both a continuous path and the set of tasks visited along it. The resulting tree therefore serves as an implicit representation of the relevant task-transition graph.

% \begin{figure}
%     \centering
%     \includegraphics[width=\linewidth]{imgs/problem.jpg}
%     \caption{Candidate tasks are colored by reward. The objective is to compute a collision-free closed-loop mission (dark green) that maximizes the collected reward under a limited travel budget. The start state is represented as the hollow gray circle. Candidate tasks outlined with green are considered collected, red are infeasible and gray are unreached by the tree.}
%     \label{fig:introduction}
% \end{figure}

This formulation offers a natural advantage over roadmap-based approaches, such as the Probabilistic Roadmap Method (PRM), for single-mission budgeted planning \cite{Penicka2019Physical}. PRM is well suited to multi-query motion planning, but its roadmap construction generally separates the exploration of the state space from the subsequent task-selection process. In contrast, RRT-based approaches grow directly from the current mission state and can incorporate motion feasibility and constraints during expansion. Collision avoidance, vehicle kinematics, nonholonomic motion, and other local feasibility conditions can be directly enforced when generating each tree edge \cite{lavalle2006planning}. Furthermore, the tree can maintain the accumulated travel cost and explicitly account for the cost required to reach a prescribed terminal state or return to the initial state. This enables task selection and motion feasibility to be evaluated within a unified incremental planning process.

Here we propose the \emph{Reward-Guided RRT*}, a sampling-based approach for budgeted task and motion planning with predefined rewards. Rewards are associated with specific task locations and are collected only when those locations are visited. Tree expansion is biased toward promising unvisited tasks, while the accumulated travel cost and terminal feasibility are explicitly considered. RRT*~\cite{karaman2011sampling} rewiring is used to improve the motion cost of promising branches, allowing the planner to discover increasingly efficient paths and potentially visit additional reward-bearing locations within the available budget. The approach can support both fixed-goal missions and closed-circuit missions in which the vehicle must return to its initial state.

%%%%%%%%%%%%%%%%%%%%%%%%%%%%%%%%%%%%%%%%%%%%%%%%%%%
\section{Related Work}
\label{sec:related_work}

% \douglas{Como esse problema já foi resolvido na literatura?}
% \douglas{O que estamos fazendo de diferente?}

The Orienteering Problem (OP) is a classical NP-hard combinatorial optimization problem that combines the selection and sequencing of locations under a limited travel budget. Given a set of reward-bearing vertices, the goal is to determine a path between prescribed endpoints that visits a subset of these vertices and maximizes the collected reward without exceeding the available budget. It was formally introduced by Golden~\etal~\cite{golden1987}, motivating the development of both exact and heuristic solution methods. Comprehensive surveys of the OP and its variants are provided by Vansteenwegen~\etal~\cite{Vansteenwegen2011} and Gunawan~\etal~\cite{gunawan2016}.

A broad range of methods has been proposed to solve the OP, including exact mixed-integer programming formulations, dynamic programming, constructive heuristics, local search, and metaheuristics \cite{gunawan2016}. These developments have led to numerous generalizations involving multiple agents~\cite{Chao1996}, time-varying~\cite{Ma2017SpatioTemporal} or spatial-correlated~\cite{Yu2016COP} rewards, clustered targets~\cite{Almeida2025Clustered}, and other application-specific constraints~\cite{Santos2021Anytime,Mansfield2022Energy,Tristao2025Team}. Such variants demonstrate the flexibility of the OP framework, but most formulations still assume that the travel cost between candidate locations is known through a predefined graph or distance matrix.

Meanwhile, in Robotics, motion planning addresses the problem of finding feasible paths for a robot operating in a continuous state space while satisfying environmental and vehicle constraints. Classical approaches include graph-search methods, potential-field methods, sampling-based methods, and path optimization techniques. Among sampling-based methods, Probabilistic Roadmaps (PRMs) construct a graph of collision-free states and feasible local connections, while Rapidly-exploring Random Trees~(RRTs) incrementally explore the state space from an initial state~\cite{Kavraki1996Probabilistic,lavalle1998rapidly}. The RRT* algorithm introduced by Karaman and Frazzoli~\cite{karaman2011sampling} extends this framework with rewiring operations and provides asymptotic optimality under suitable conditions. These methods are particularly attractive for high-dimensional spaces and problems involving complex geometric or kinodynamic constraints.

% \textcolor{red}{
% The combination of reward-based task selection and motion planning has received increasing attention in robotics. The Physical Orienteering Problem (POP), introduced by Pěnička~\etal~\cite{Penicka2019Physical}, explicitly integrates the OP with motion planning in environments containing obstacles. In the POP, a robot must visit a subset of predefined targets to maximize collected reward while following a feasible collision-free path within a limited travel budget. The authors solve the resulting problem by combining a high-level OP solver with motion-planning techniques, highlighting the fundamental coupling between combinatorial task selection and continuous motion feasibility.}

% \textcolor{red}{
% Other formulations have incorporated increasingly complex vehicle and environmental constraints. The Dubins Orienteering Problem considers vehicles with curvature constraints and therefore requires the selected route to satisfy a minimum turning-radius constraint~\cite{Penicka2017Dubins}. Similarly, the Minimal Exposure Dubins Orienteering Problem extends this setting by jointly considering reward collection and exposure to a known sensor field~\cite{Macharet2021Minimal}. These works demonstrate that realistic robotic orienteering cannot always be adequately represented by Euclidean distances between targets, since the feasibility and cost of traveling between locations depend on the vehicle model and environment.
% }

The integration of reward-based task selection and motion planning has received increasing attention, motivated by the need to account for environmental and vehicle constraints that cannot be captured by simple distances between reward locations. The Physical Orienteering Problem~(POP), introduced by Pěnička~\etal~\cite{Penicka2019Physical}, extends the classical OP to environments with obstacles, requiring selected targets to be connected by feasible collision-free trajectories. The Dubins Orienteering Problem (DOP) further incorporates curvature constraints into the vehicle model~\cite{Penicka2017Dubins}, while the Minimal Exposure Dubins Orienteering Problem considers both reward collection and exposure to a known sensor field~\cite{Macharet2021Minimal}. More recently, the Kinematic Orienteering Problem (KOP) integrates reward-maximizing node selection with time-optimal trajectory generation under velocity and acceleration constraints for multirotor UAVs~\cite{Meyer2022Kinematic}. 

These works illustrate the broad range of environmental objectives, vehicle models, and motion constraints that can be incorporated into robotic orienteering. However, although the literature establishes a clear connection between the OP and motion planning, existing approaches commonly retain a decomposition in which candidate task locations and their transitions are first represented in a discrete problem, while motion planning is subsequently used to validate or realize the selected route. 
In contrast, our proposed approach uses RRT*-based trees to incrementally explore the continuous state space while simultaneously guiding the search toward rewarding task locations. Rather than introducing another problem-specific variant of the Orienteering Problem, we propose a general sampling-based planning framework that jointly accounts for task selection, path feasibility, motion constraints, and budget requirements.

%%%%%%%%%%%%%%%%%%%%%%%%%%%%%%%%%%%%%%%%%%%%%%%%%%%%
\section{Problem Formulation}
\label{sec:problem_formulation}

% \douglas{Definição formal do problema.}
% \douglas{Falar das principais características e restrições.}

We consider a vehicle operating in bounded environments with obstacles and a set of reward-bearing locations. Our goal is to compute a feasible path that maximizes the total reward collected by visiting a subset of these locations while satisfying a limited travel budget. The problem can be viewed as a motion-constrained extension of the Orienteering Problem, in which the reward-bearing locations are known, but the feasible paths and their associated costs are not explicitly available beforehand.

Let the environment be represented by a state space $\mathcal{X}$ and its collision-free subset $\mathcal{X}_{\mathrm{free}} \subseteq \mathcal{X}$. Let
$\mathcal{V} = \{v_1,v_2,\ldots,v_N\}$ denote the set of reward-bearing locations, where each location $v_i \in \mathcal{X}_{\mathrm{free}}$ is associated with a non-negative reward $r_i$. The vehicle starts at position $\stateinit$ and must reach a terminal position $\stategoal$. The terminal position may be distinct from the initial position or may coincide with it, corresponding to a closed-circuit mission.

A possible solution is a continuous path $\sigma : [0,1] \rightarrow \mathcal{X}_{\mathrm{free}}$ satisfying the vehicle's motion constraints, with $\sigma(0)=\stateinit$ and $\sigma(1)=\stategoal$. Also, let $c(\sigma)$ denote the cost of this path, typically its length or travel time, and let $B$ denote the available budget, such that $\sigma$ must satisfy $c(\sigma) \leq B$.

{

Now, let $r_g$ denote the visitation radius associated with $v_i$. A reward location is considered visited if the path enters its corresponding visitation region. Accordingly, the set of locations visited by $\sigma$ is defined as
\begin{equation} 
\mathcal{V}(\sigma)
=
\left\{
v_i \in \mathcal{V}
~|~
\min_{t\in[0,1]}
d\left(\sigma(t),v_i\right)
\leq r_g
\right\},
\nonumber
\end{equation}
where $d(\cdot,\cdot)$ is an appropriate distance metric in the state space.
}
The goal is to maximize the total collected reward:
\begin{equation*} 
\max_{\sigma}
\quad
R(\sigma)
=
\sum_{v_i \in \mathcal{V}(\sigma)} r_i, ~~~\text{subject to}
\end{equation*}
%
% subject to
%
\begin{align} 
\sigma(t) & \in \mathcal{X}_{\mathrm{free}}, \nonumber
\quad \forall t \in [0,1], \\ \nonumber
c(\sigma) &\leq B, \\ \nonumber
\sigma(0) &=\stateinit,
\quad
\sigma(1)=\stategoal.
\nonumber
\end{align}

Unlike the classical graph-based formulation of the OP, the cost of traveling between two reward locations is not assumed to be known. In particular, for two locations $v_i$ and $v_j$, the existence and cost of a feasible connection depend on the environment and the vehicle's motion constraints. Thus, the underlying task-transition graph is not explicitly provided. A feasible transition can be represented by a motion-planning problem,
\begin{equation} 
\sigma_{ij}^{*}
=
\arg \min_{\sigma} c(\sigma), ~~~\text{subject to}
\nonumber
\end{equation}
%
% subject to
%
\begin{equation} 
\sigma(0) =v_i,
\quad
\sigma(1)=v_j,
\quad
\sigma(t)\in\mathcal{X}_{\mathrm{free}},
\nonumber
\end{equation}
together with the vehicle-specific motion constraints. However, computing such connections for all pairs of reward locations would require constructing the complete underlying graph, which may be computationally expensive and unnecessary for a single budget-constrained mission.

The central challenge addressed in this work is to jointly determine the subset and ordering of reward locations to visit while constructing feasible connections between them. Rather than assuming a precomputed transition graph, the proposed method incrementally explores the continuous state space and discovers relevant feasible task transitions during planning. The resulting search must balance three competing objectives: collecting additional reward, maintaining motion feasibility, and preserving sufficient budget to reach the required terminal position.

%%%%%%%%%%%%%%%%%%%%%%%%%%%%%%%%%%%%%%%%%%%%%%%%%%%%
\section{Methodology}
\label{sec:methodology}

%%%%%%%%%%%%%%%%%%%%%%%%%%%%%%%%%%%%%%%%%%%%%%%%%%%%
\subsection{Algorithm Overview}
\label{sec:algorithm_overview}

The proposed method constructs a budget-feasible mission incrementally through a sequence of reward-guided RRT* searches. At each stage, an outgoing tree $\mathcal{T}_{\mathrm{out}}$ is rooted at the endpoint of the currently confirmed mission and explores the collision-free state space in search of an unvisited reward-bearing location.
When the outgoing tree reaches a candidate location, the corresponding path is not immediately incorporated into the mission. Instead, a return tree $\mathcal{T}_{\mathrm{ret}}$ is initialized at the candidate state and attempts to connect it to the prescribed terminal state, which in the closed-circuit case coincides with $\stateinit$. The candidate is committed only if the complete outgoing-and-return path satisfies the available travel budget.

\begin{figure*}[t]
    \centering
    \includegraphics[width=.85\linewidth]{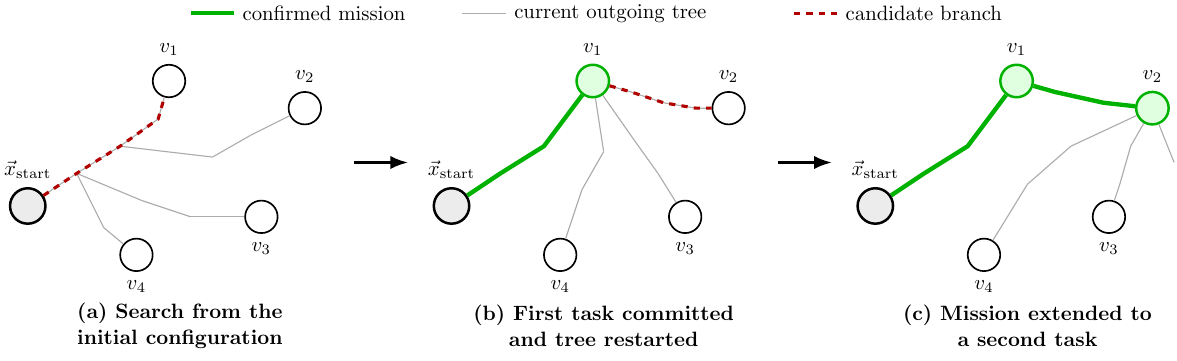}
    \caption{
    The outgoing tree first searches from the initial state and reaches task $v_1$. After the candidate is validated, its branch is \emph{committed} to the route and the next outgoing tree is rooted at $v_1$. The same procedure is repeated when $v_2$ is confirmed. Gray edges denote the current search tree, while green edges denote the already committed ones.
    }
    \label{fig:incremental_mission}
    \vspace{-4mm}
\end{figure*}

After a successful validation, the outgoing segment is appended to the confirmed mission, the candidate reward is collected, and a new outgoing tree is initialized at the newly visited location. The new tree inherits the travel cost and reward accumulated by the mission. Therefore, the planner does not construct a single tree containing all possible task sequences. Instead, it generates a sequence of local trees that incrementally extend one feasible mission.
Figure~\ref{fig:incremental_mission} illustrates this process. The solid green path represents the mission already committed to by the planner, whereas the gray branches represent the current outgoing search tree. Once a candidate is validated, the corresponding branch becomes part of the confirmed mission, and the next tree is rooted at that location.

Algorithm~\ref{alg:mission_planner} summarizes the high-level procedure, while Algorithms~\ref{alg:outgoing_expansion} and~\ref{alg:return_validation} detail the outgoing-tree expansion and return-validation routines, respectively.

\begin{algorithm}[htpb]
\caption{Reward-Guided Incremental Planning}
\label{alg:mission_planner}

\KwIn{
    state $\stateinit$,
    reward locations $\mathcal{V}$,
    budget $B$
}

\KwOut{
    confirmed mission $\mathcal{P}$
}

$\mathcal{P} \gets [\stateinit]$, %\; 
$\mathcal{V}_{\mathrm{visited}} \gets \emptyset$, %\;
$\mathcal{V}_{\mathrm{failed}} \gets \emptyset$\;

$\stateroot \gets \stateinit$\;
$C_{\mathrm{root}} \gets 0$, %\;
$R_{\mathrm{root}} \gets 0$\;

$\mathcal{T}_{\mathrm{out}}
\gets
\texttt{InitializeTree}
(\stateroot,C_{\mathrm{root}},R_{\mathrm{root}})$\;

\While{\texttt{PlanningActive}()}{

    $(g,\state_g,\mathrm{status})
    \gets
    \texttt{GrowOutgoingTree}
    \left(
        \mathcal{T}_{\mathrm{out}},
        \mathcal{V}_{\mathrm{visited}},
        \mathcal{V}_{\mathrm{failed}},
        B
    \right)$\;
    
    \If{$\mathrm{status}=\mathrm{NoCandidate}$}{
        \textbf{break}\;
    }

    $(\mathrm{success},
      \mathcal{P}_{g},
      \mathcal{P}_{\mathrm{ret}})
    \gets
    \texttt{ValidateReturn}
    \left(
        \mathcal{T}_{\mathrm{out}},
        \state_g,
        \stategoal,
        B
    \right)$\;
    
    \eIf{$\mathrm{success}$}{

        $\mathcal{P}
        \gets
        \texttt{CommitSegment}
        (\mathcal{P},\mathcal{P}_{g})$\;

        $\mathcal{V}_{\mathrm{visited}}
        \gets
        \mathcal{V}_{\mathrm{visited}}\cup\{g\}$\;

        $(\mathcal{P},\mathcal{P}_{\mathrm{ret}})
        \gets
        \texttt{RefinePaths}
        (\mathcal{P},\mathcal{P}_{\mathrm{ret}})$\;

        $\stateroot \gets \state_g$\;
        $C_{\mathrm{root}} \gets
        \texttt{Cost}(\mathcal{P})$, %\;
        $R_{\mathrm{root}} \gets
        \displaystyle
        \sum_{v_i\in\mathcal{V}_{\mathrm{visited}}} r_i$\;

        $\mathcal{T}_{\mathrm{out}}
        \gets
        \texttt{InitializeTree}
        (\stateroot,C_{\mathrm{root}},R_{\mathrm{root}})$\;

    }{

        $\mathcal{V}_{\mathrm{failed}}
        \gets
        \mathcal{V}_{\mathrm{failed}}\cup\{g\}$\;

    }
}

\Return $\mathcal{P}$\;

\end{algorithm}

%%%%%%%%%%%%%%%%%%%%%%%%%%%%%%%%%%%%%%%%%%%%%%%%%%%%
\subsection{Outgoing Tree Expansion}
\label{sec:outgoing_tree}

The outgoing tree is responsible for discovering the next task to be incorporated into the mission. At each planning stage, a new tree is rooted at the last confirmed task (or at the initial state during the first iteration) and expanded until either a new candidate task is reached or no additional feasible candidates can be identified.

Unlike conventional graph-based formulations of the Orienteering Problem (OP), the proposed method explores the continuous collision-free state space by incrementally growing an RRT*. Each expansion follows the standard sampling-based planning procedure, consisting of sampling, nearest-neighbor search, steering, parent selection, and rewiring \cite{karaman2011sampling}. The main adaptations are (i) a reward-guided sampling strategy that biases the exploration toward promising task locations, (ii) a budget feasibility test that prevents the generation of infeasible branches, and (iii) the detection of candidate tasks that trigger the return-validation phase. Note that because each tree is grown for a finite iteration budget 
$I$ and is subsequently frozen once a task is committed, the classical asymptotic-optimality guarantee of RRT*~\cite{karaman2011sampling} does not directly apply to the overall mission; rewiring here serves as an anytime cost-improvement mechanism within each stage's finite budget, not as a source of global optimality guarantees.
Algorithm~\ref{alg:outgoing_expansion} summarizes one iteration of the outgoing-tree expansion. The following subsections describe the components that differ from the standard RRT*.

\begin{algorithm}[htpb]
\caption{\texttt{GrowOutgoingTree}($\cdot$)}
\label{alg:outgoing_expansion}

\KwIn{
outgoing tree $\mathcal{T}$,
visited tasks $\mathcal{V}_{visited}$,
failed tasks $\mathcal{V}_{failed}$,
budget $B$
}
\KwOut{
candidate task $g$,
candidate state $\state_g$,
status
}

\While{planning budget available}{

    $\staterand\gets
    \texttt{RewardGuidedSample}
    (\mathcal{V}\setminus
    (\mathcal{V}_{visited}\cup
    \mathcal{V}_{failed}))$\;

    $\statenearest\gets
    \texttt{Nearest}
    (\mathcal{T},\staterand)$\;

    $\statenew \gets
    \texttt{Steer}
    (\statenearest,\staterand)$\;

    \If{\texttt{ObstacleFree}
    $(\statenearest,\statenew)$}{

        $\mathcal{X}_{near}
        \gets
        \texttt{Near}
        (\mathcal{T},\statenew)$\;

        $\state_{parent}
        \gets
        \texttt{ChooseParent}
        (\mathcal{X}_{near},\statenew)$\;

        \If{\texttt{BudgetFeasible}
        $(\state_{parent},\statenew,B)$}{

            $\texttt{AddVertex}
            (\mathcal{T},\statenew)$\;

            $\texttt{AddEdge}
            (\mathcal{T},\state_{parent},\statenew)$\;

            $\texttt{Rewire}
            (\mathcal{T},\statenew,\mathcal{X}_{near})$\;

            \If{\texttt{TaskReached}
            $(\statenew)$}{

                % \Return
                % candidate task\;
                $g \gets \texttt{ReachedTask}(\statenew)$\;

                $\state_g \gets \statenew$\;

                \Return
                $(g,\state_g, \textrm{Candidate})$\;

            }

        }

    }

}

% \Return
% NoCandidate\;
\Return $(\varnothing,\varnothing, \textrm{NoCandidate})$\;

\end{algorithm}

%%%%%%%%%%%%%%%%%%%%%%%%%%%%%%%%%%%%%%%%%%%%%%%%%%%%
\subsubsection{Reward-Guided Sampling}

To balance exploration and exploitation, the sampling strategy combines uniform random exploration with goal-directed sampling.
With probability $p_r$, the algorithm samples a random state from the collision-free workspace, preserving the exploratory behavior of the original RRT*. Otherwise, the algorithm selects one of the unvisited task locations according to a roulette-wheel strategy.

Let $r_i$ denote the reward associated with task $v_i$ and let $d_i$ represent its Euclidean distance to the root of the current outgoing tree. The sampling weight assigned to $v_i$ is $w_i=r_i/d_i,$
% %
% \begin{equation*}
%     w_i=\frac{r_i}{d_i},
%     %\label{eq:roulette}
% \end{equation*}
%
which favors nearby high-reward tasks while still allowing every candidate to be selected. The probability of selecting task $v_i$ is therefore
\begin{equation}
    P(v_i)= \frac{w_i}{\sum_j w_j}.
    \label{eq:roulette_prob}
\end{equation}

This mechanism continuously biases the tree toward promising regions of the workspace without sacrificing the global exploration capability inherited from the RRT*.

%%%%%%%%%%%%%%%%%%%%%%%%%%%%%%%%%%%%%%%%%%%%%%%%%%%%
\subsubsection{Budget-Constrained Expansion}

Each node stores the cumulative travel cost from the root of the current mission. After selecting the parent node, the accumulated cost of the new state is updated as
\begin{equation}
    C(\state_{\mathrm{new}})
    =
    C(\state_{\mathrm{parent}})
    +
    c(\state_{\mathrm{parent}},\state_{\mathrm{new}}),
    \label{eq:budget}
\end{equation}
where $c(\cdot)$ denotes the Euclidean length of the newly added edge.
Nodes whose accumulated travel cost exceeds the available mission budget are immediately discarded. Consequently, every branch in the outgoing tree satisfies the travel-budget constraint, preventing the planner from expanding toward regions it cannot reach within the remaining mission budget.

%%%%%%%%%%%%%%%%%%%%%%%%%%%%%%%%%%%%%%%%%%%%%%%%%%%%
\subsubsection{Candidate Task Detection}

After inserting a new node into the tree, the planner verifies whether the new state lies inside the capture region of an unvisited task. A task $v_i$ becomes a candidate whenever
\begin{equation}
    \left\| \state_{\mathrm{new}} - v_i \right\| < r_g,
    \label{eq:goal_reached}
\end{equation}
where $r_g$ denotes the task capture radius.

Reaching a task does not immediately incorporate it into the mission. Instead, the outgoing tree is temporarily suspended, and a second RRT* is initialized from the candidate task to verify whether a feasible return to the $\stateinit$ still exists within the available travel budget. Only tasks that satisfy this validation are permanently incorporated into the mission.

%%%%%%%%%%%%%%%%%%%%%%%%%%%%%%%%%%%%%%%%%%%%%%%%%%%%
\subsection{Return Feasibility Validation}
\label{sec:return_validation}

Reaching a task with the outgoing tree does not guarantee that the corresponding task can be incorporated into the mission. Since the proposed planner constructs paths subject to a limited travel budget, every candidate task must also admit a feasible return to the goal state (Alg.~\ref{alg:return_validation}).

\begin{algorithm}[htpb]
\caption{\texttt{ValidateReturn}($\cdot$)}
\label{alg:return_validation}

\KwIn{
outgoing tree $\mathcal{T}_{out}$,
candidate state $\state_g$,
goal state $\state_\mathrm{goal}$,
remaining budget $B$
}

\KwOut{
success flag,
outgoing path $P_g$,
return $P_{ret}$
}

$\mathcal{T}_{ret}
\gets
\texttt{InitializeTree}(\state_g)$\;

\While{return planning active}{

    $\staterand\gets
    \texttt{RewardGuidedSample}
    (\stategoal)$\;

    $\statenearest
    \gets
    \texttt{Nearest}
    (\mathcal{T}_{ret},\staterand)$\;

    $\statenew
    \gets
    \texttt{Steer}
    (\statenearest,\staterand)$\;

    \If{\texttt{ObstacleFree}
    $(\statenearest,\statenew)$}{

        $\mathcal{X}_{near}
        \gets
        \texttt{Near}
        (\mathcal{T}_{ret},\statenew)$\;

        $\state_{parent}
        \gets
        \texttt{ChooseParent}
        (\mathcal{X}_{near},\statenew)$\;

        \If{\texttt{BudgetFeasible}
        $(\state_{parent},\statenew,B)$}{

            $\texttt{AddVertex}
            (\mathcal{T}_{ret},\statenew)$\;

            $\texttt{AddEdge}
            (\mathcal{T}_{ret},\state_{parent},\statenew)$\;

            $\texttt{Rewire}
            (\mathcal{T}_{ret},\statenew,\mathcal{X}_{near})$\;

            \If{\texttt{ReachedGoal}
            $(\statenew)$}{

                $P_g \gets
                \texttt{ExtractPath}(\mathcal{T}_{out},\state_g)$\;

                $P_{ret} \gets
                \texttt{ExtractPath}(\mathcal{T}_{ret},\statenew)$\;

                \Return
                $(\texttt{True},P_g,P_{ret})$\;

            }

        }

    }

}

\Return
$(\texttt{False},\texttt{None},\texttt{None})$\;

\end{algorithm}

To verify this condition, the planner initializes a second RRT* rooted at the candidate task and expands it toward the initial state. The return tree follows the same expansion procedure adopted by the outgoing planner, including nearest-neighbor search, steering, parent selection, and rewiring. However, unlike the outgoing phase, the reward-guided sampling is performed using only the goal state, because the objective is no longer to discover additional tasks, but solely to determine the feasibility of the return path.

The accumulated travel cost stored in the root of the return tree corresponds to the travel cost already incurred by the outgoing mission. Consequently, each node in the return tree represents the total travel cost of completing the mission through that partial return path. Nodes exceeding the available mission budget are immediately discarded.
Whenever the return tree reaches the initial state, the candidate task is considered feasible. The outgoing path segment leading to the task is permanently incorporated into the mission, the return path is stored as the current feasible terminal path, and the planner initializes a new outgoing tree rooted at the newly confirmed task.

Conversely, if the return tree fails to connect the candidate task to the goal state within the prescribed budget, the task is temporarily classified as unavailable until a new location is selected. The planner then resumes the outgoing-tree exploration in search of alternative task locations.

%%%%%%%%%%%%%%%%%%%%%%%%%%%%%%%%%%%%%%%%%%%%%%%%%%%%
\subsection{Path Refinement}
\label{sec:path_refinement}

The paths generated by the outgoing and return trees are feasible but not necessarily locally optimal. Since both trees are constructed incrementally, the resulting paths may contain unnecessary intermediate nodes and detours introduced by the sampling process. Therefore, after each successful task insertion, a post-processing stage reduces the overall travel distance before the next planning stage begins.

The first refinement consists of a greedy shortcut procedure \cite{lavalle2006planning}. Starting from the end of each path segment, the algorithm iteratively attempts to replace a sequence of consecutive edges with a single straight-line connection. A shortcut is accepted whenever the resulting segment is collision-free and reduces the accumulated travel distance. The procedure is repeated until no additional local improvements can be found. Since only collision-free connections are accepted, path feasibility is always preserved.

After the shortcut stage, a 2-opt local optimization is applied to the complete mission \cite{croes1958method}. The algorithm evaluates alternative edge pairings by reversing intermediate path segments whenever such a modification reduces the total travel distance while maintaining collision-free connections. Unlike the greedy shortcut, which removes unnecessary intermediate nodes, the 2-opt operator optimizes the ordering of consecutive path segments and eliminates local geometric inefficiencies that remain after the shortcut stage.

Both refinement procedures operate exclusively on the geometry of the confirmed mission and do not modify the set of visited tasks. Consequently, the collected reward remains unchanged while the resulting path becomes shorter, increasing the remaining travel budget available for subsequent planning stages.

%%%%%%%%%%%%%%%%%%%%%%%%%%%%%%%%%%%%%%%%%%%%%%%%%%%%
\section{Experiments}

% \douglas{Como iremos avaliar o desempenho?}
% \douglas{É possível comparar com algum baseline?}

\subsection{Illustrative Examples}

% To evaluate the proposed motion planning Reward-Guided RRT*, it was tested on three different scenery of distinct difficulties and distinct sets of reward-bearing locations (which will be referred as goals), but all sharing the same initial and final position of our path, making it a closed-circuit mission. The difficulties were separated as: LOW, MEDIUM and HIGH; Indicating the increase of obstacle complexity and number of goals. The position and reward-values of the goals were randomly generated.

\begin{figure*}[t]
     \centering
     \begin{subfigure}[b]{0.325\linewidth}
         \centering
         \includegraphics[width=\linewidth]{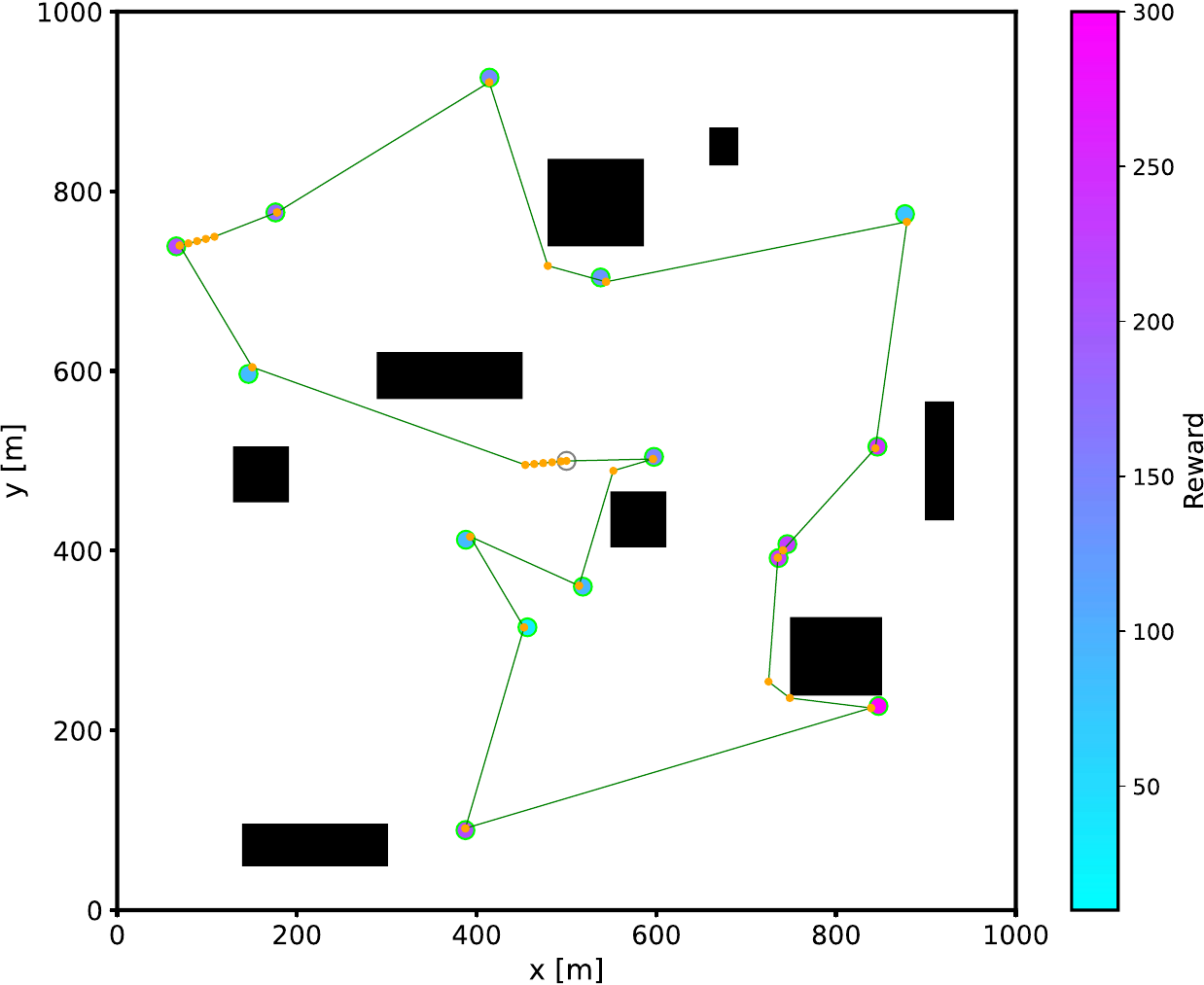}
         \caption{LOW}
     \end{subfigure}
     \begin{subfigure}[b]{0.325\linewidth}
         \centering
         \includegraphics[width=\linewidth]{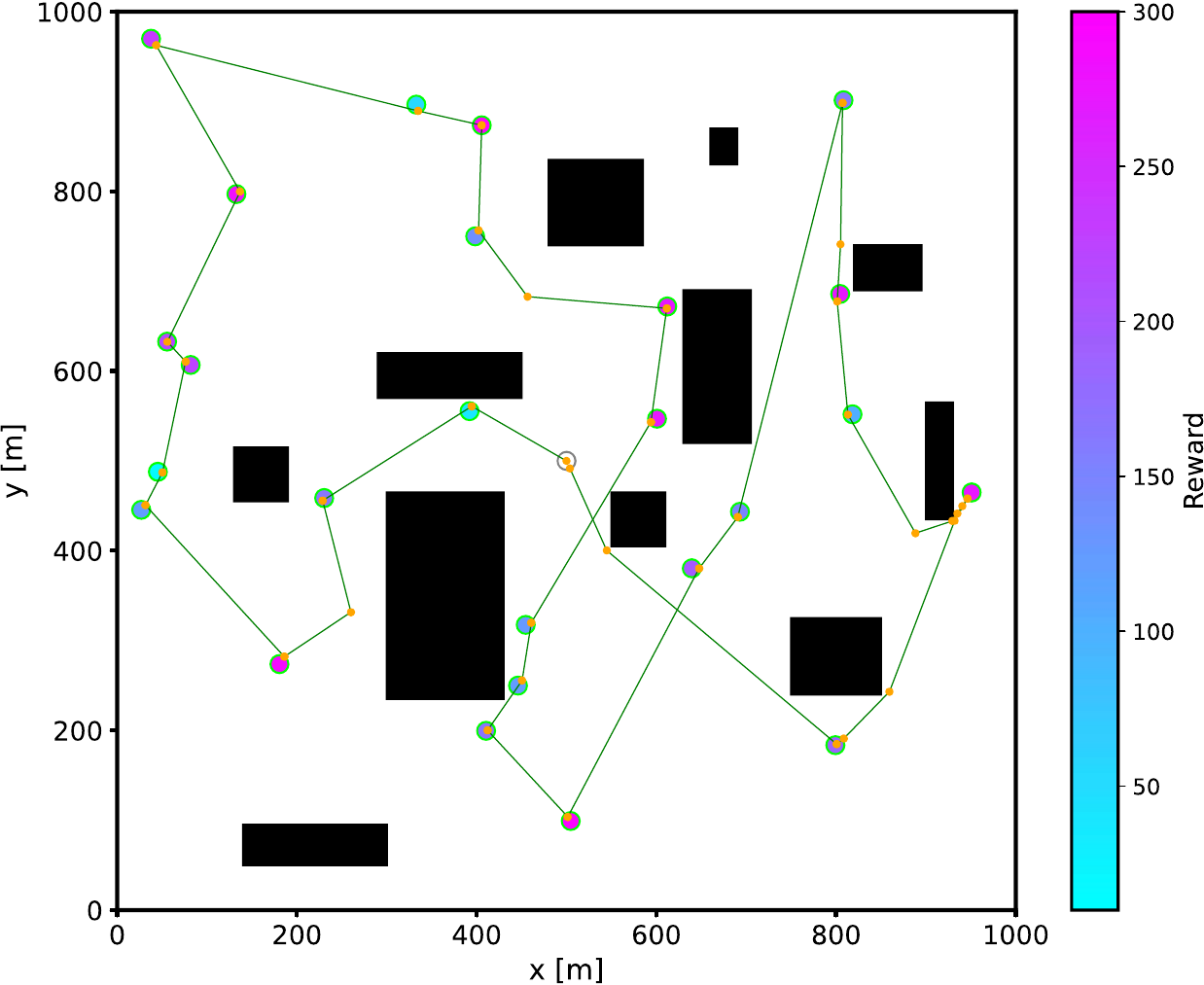}
         \caption{MEDIUM}
     \end{subfigure}
     \begin{subfigure}[b]{0.325\linewidth}
         \centering
         \includegraphics[width=\linewidth]{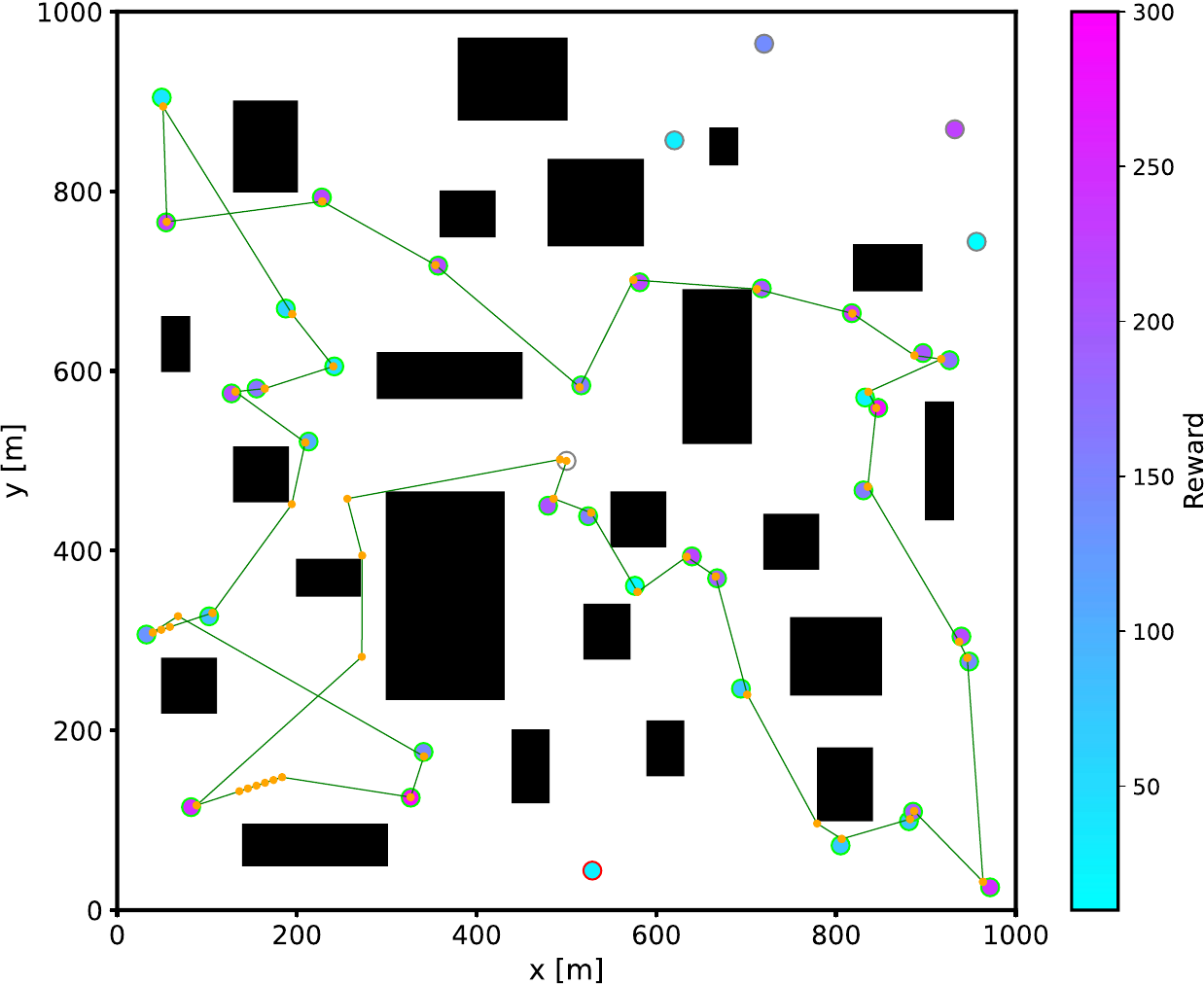}
         \caption{HIGH}
     \end{subfigure}
        \caption{Example solutions for the (a) LOW, (b) MEDIUM, and (c) HIGH scenarios, with 15, 25, and 40 goals, respectively. All results use $I=25000$ and $d=10\mathrm{m}$; each result shows the best run (maximum reward) obtained.}
        \label{fig:example-results}
\end{figure*}

To evaluate our Reward-Guided RRT* motion planning framework, we consider three $1000\mathrm{m}\times1000\mathrm{m}$ scenarios with increasing difficulty and different sets of reward-bearing locations, hereafter referred to as goals. All scenarios share the same initial and final position, $[500,500]$, resulting in a closed-circuit mission. We categorize them as LOW, MEDIUM, and HIGH, reflecting increasing obstacle density and complexity, as well as an increasing number of goals. Task positions and reward values are randomly generated for each scenario. Figure~\ref{fig:example-results} shows the best solution over 30 runs for each scenario, defined as the solution with the highest reward and, among those, the lowest budget.

% \begin{figure}
%     \centering
%     \includegraphics[width=\linewidth]{imgs/best_LOW_run/path_4.pdf}
%     \caption{Close-circuit path solution of LOW difficult scenery with $I=25000$ and $d=10m$.}
%     %Colocar a porcentagem do reward e budget%
%     \label{fig:introduction}
% \end{figure}
% \begin{figure}
%     \centering
%     \includegraphics[width=\linewidth]{imgs/best_MEDIUM_run/path_1.pdf}
%     \caption{Close-circuit path solution of MEDIUM difficult scenery with $I=25000$ and $d=10m$.}
%     \label{fig:introduction}
% \end{figure}
% \begin{figure}
%     \centering
%     \includegraphics[width=\linewidth]{imgs/best_HIGH_run/path_15.pdf}
%     \caption{Close-circuit path solution of HIGH difficult scenery with $I_{max}=25000$ and $d=10m$.}
%     \label{fig:introduction}
% \end{figure}

% \douglas{Descrever cenários e mostrar exemplos de rotas resultantes.}

\subsection{Quantitative Analysis}

% Sets of 30 runs of testing varying the combination of: max tree expansion iterations $I$, steer step distance $d$ and scenery difficult, making groups of: (1500, 5000, 10000, 15000, 25000, 5000), (10, 25, 50) and (LOW, MEDIUM, HIGH), respectively; were made using this paper approach. Figure 3 show the final solution and the best run (highest reward collected and less budget used) found for each scenery.

% Statistical analyzes of the mean budget used $B\%$ and the mean reward $R\%$ demonstrate that our method has a low result variation while being non-deterministic. It also show a expected behavior that with the increase of goals and obstacle complexity, a higher number of iterations is necessary to achieve a better result and make the most of the budget available, while a bigger expansion step can help with low max iterations.

% Interesting enough, the graph shows that past 10000 iterations the results stabilize and have little to none improvement for budget or reward regardless of the step distance.

% Due to its non-deterministic nature, results outside the stability point can, but not necessarily, still contain the best solution, as all the best solutions found during experiments are beyond this point. Also an expected behavior, since a low expansion step of a RRT while having a high enough maximum number of iterations causes much more sophisticated and precise paths while increasing the computational time, thus, the stability point must be used as metric for when is no longer worth increasing the computational time.

\begin{figure*}[t]
    \centering
    \includegraphics[width=\linewidth]{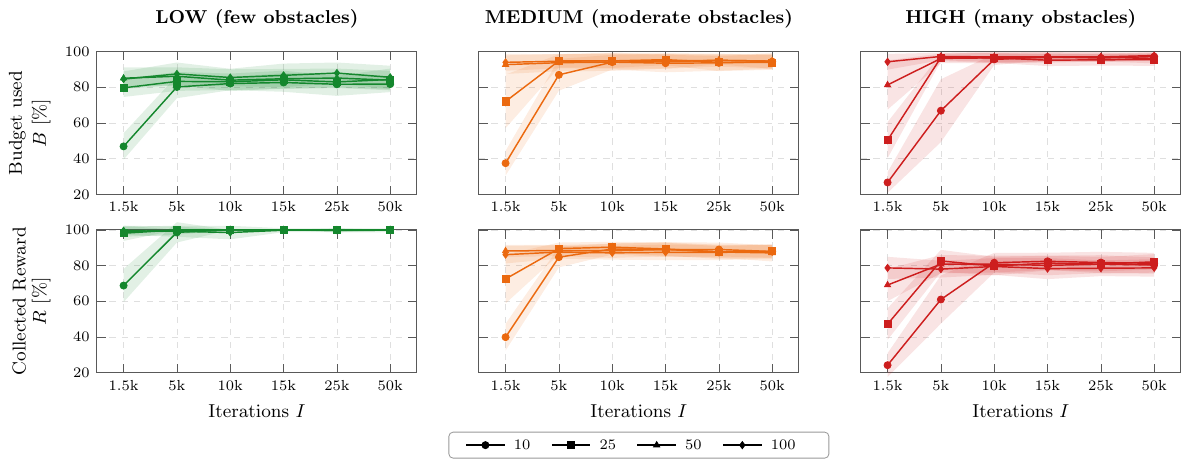}
    \caption{
    Overall statistical analysis of the proposed Reward-Guided RRT*. Each curve represents the mean over 30 runs for a fixed steering step $d$, with different steps indicated by distinct markers. Shaded regions represent one standard deviation. %The dotted horizontal line indicates the near-optimal result obtained with VNS-PRM*~\cite{Penicka2019Physical}.
    }
    \label{fig:statistical}
\end{figure*}

To quantitatively evaluate the performance and stability of our proposed approach, we conducted a systematic set of experiments, varying the main planning parameters and scenario difficulty.
For each state, we ran 30 independent trials, varying the maximum number of tree expansion iterations $I$, the steering step distance $d$, and the scenario difficulty. The evaluated values were $I \in \{1500, 5000, 10000, 15000, 25000\}$, $d \in \{10,25,50,100\}$, and difficulty levels LOW, MEDIUM, and HIGH. Figure~\ref{fig:statistical} summarizes the overall results and these trends.

Statistical analysis of the mean budget utilization $B\%$ and mean collected reward $R\%$ shows relatively low variability across runs while preserving the method's non-deterministic behavior. The results also show the expected behavior: more challenging scenarios, characterized by more goals and increased obstacle complexity, generally require more iterations to obtain higher-quality solutions and better exploit the available budget. Conversely, a larger steering step can improve performance when the maximum number of iterations is limited, as it lets the tree explore the environment more quickly.

Interestingly, the results indicate that, beyond approximately 10,000 iterations, the performance tends to stabilize, with little to no further improvement in either budget utilization or collected reward, regardless of the steering step distance. This suggests the existence of a practical stability point beyond which increasing the number of iterations provides diminishing returns.
Due to the non-deterministic nature of the sampling-based approach, however, solutions of higher quality may still occasionally be obtained beyond this stability point. In fact, all of the best solutions observed across the experiments were obtained in runs beyond this point. This behavior is consistent with the trade-off inherent to RRT-based planning: smaller steering steps with sufficiently large iteration budgets allow the tree to explore the state space more finely, potentially producing more sophisticated and precise paths, but at the cost of increased computational time. Therefore, the observed stability point provides a practical criterion for determining when further increases in computational effort are unlikely to yield significant improvements in solution quality.

% \douglas{Colocar gráficos e fazer discussão.}

\subsection{State-of-the-art Comparison}

% In order to see the performance of the Reward-Guided RRT* compared to another path-finding algorithm, the VNS-PRM* [3] algorithm was run in each proposed experimental scenario difficult. Considering that the final path given by the VNS-PRM* is the best it can get, to make a fair comparison, the VNS-PRM* results were compared with the best results obtained with the Reward-Guided RRT* for each scenario, as shown in Table~\ref{tab:comparaca_caminhos}. As It was stated, all the best results taken from the Reward-Guided RRT* came from the 25000 max tree expansion iterations with a step distance value of 10 combination. The best routes obtained with the Reward-Guided RRT* for all scenarios can be seen in Figure~\ref{fig:example-results}.

To assess the performance of our Reward-Guided RRT*, we compare it with VNS-PRM*~\cite{Penicka2019Physical}, which represents the closest approach to our method. Both techniques address the integration of reward-based route selection and motion planning in environments with obstacles using sampling-based strategies. However, they differ fundamentally in how they use sampling. The VNS-PRM* relies on a classical Probabilistic Roadmap to build a dense representation of the free state space, while our approach incrementally grows an RRT* tree from the current mission state. In particular, the classical PRM formulation does not directly incorporate motion constraints into the sampling and connection process as an RRT-based planner can, allowing our approach to naturally account for motion feasibility during tree expansion in Sec.~\ref{sec:kinematic}.

For a fair comparison, we evaluated VNS-PRM* in each proposed scenario and compared its resulting solutions with the best solutions obtained by Reward-Guided RRT*, as reported in Table~\ref{tab:comparaca_caminhos}. The best solutions obtained with Reward-Guided RRT* used $I=25000$ iterations and a steering step of $d = 10$ (Figure~\ref{fig:example-results}).

\begin{table}[htbp]
\centering
\caption{Comparison of the best results obtained with Reward-Guided RRT* and VNS-PRM*~\cite{Penicka2019Physical}.}
\label{tab:comparaca_caminhos}
\begin{tabular}{c cccc}
\toprule
 & \multicolumn{2}{c}{Reward-Guided RRT*} & \multicolumn{2}{c}{VNS-PRM*~\cite{Penicka2019Physical}} \\
\cmidrule(lr){2-3} \cmidrule(lr){4-5}
 & $\% {B}$ & $\% {R}$ & $\% {B}$ & $\% {R}$ \\
\midrule
LOW    & 0.690 & 1 & 0.668 & 1 \\
MEDIUM & 0.986 & 1 & 0.896 & 1 \\
HIGH   & 0.966 & 0.929 & 0.987 & 0.967 \\
\bottomrule
\end{tabular}

\end{table}

% The results showed that the VNS-PRM* always got the better solution but the Reward-Guided RRT* were not too far from it. Other important analysis that should not be left out is how big is the number of samples the VNS-PRM* algorithm needs to get the best path compared to the number of nodes the Reward-Guided RRT* produces to get to the final path, which is something that makes the Reward-Guided RRT* have an advantage in this aspect. A visual and quantitative comparison can be made with Figure~\ref{fig:comparacao_nodes} and Table~\ref{tab:comparacao_nodes}, respectively.

\begin{figure}[htpb]
     \centering
     \begin{subfigure}[b]{\linewidth}
         \centering
          \includegraphics[trim={9mm 6mm 12mm 1mm},clip,height=4.1cm]{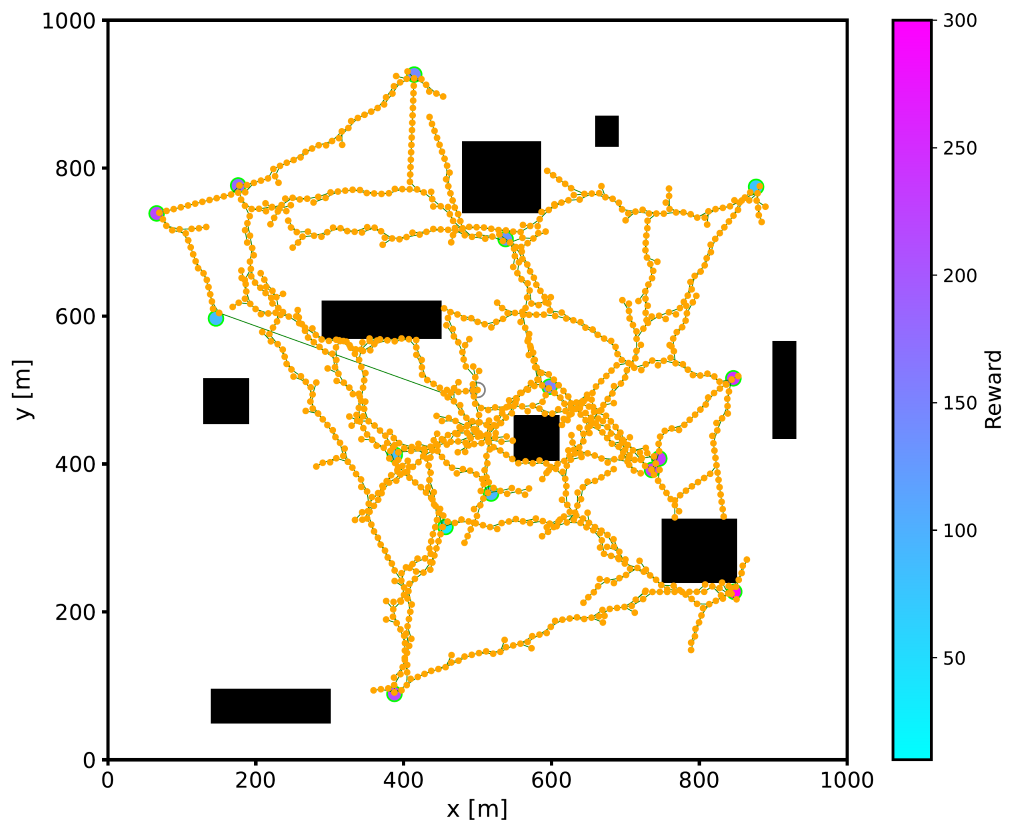}
         \includegraphics[trim={2cm 2cm 2cm 2cm},clip,height=4.1cm]{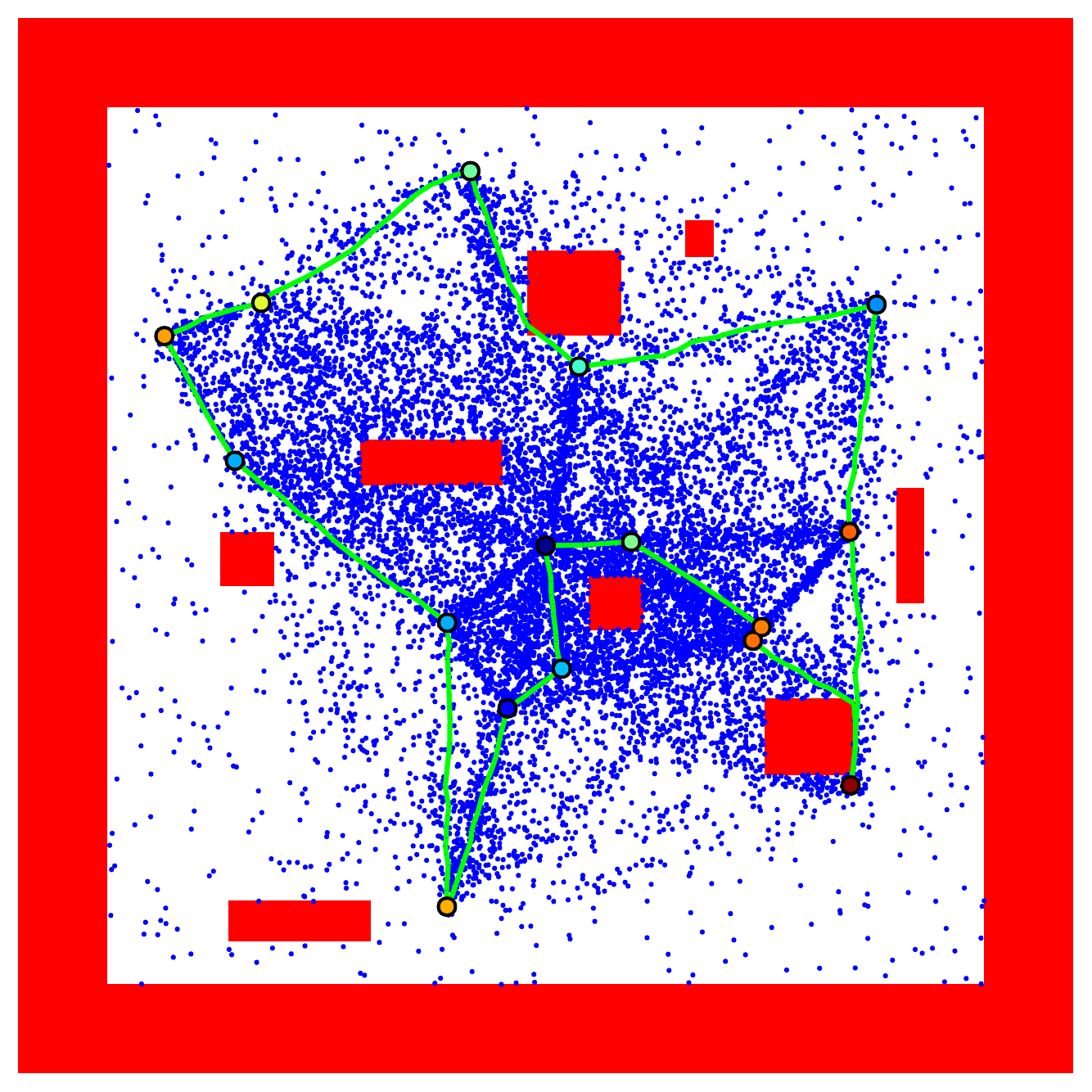}
         \caption{LOW}
     \end{subfigure}
     \begin{subfigure}[b]{\linewidth}
         \centering
         \includegraphics[trim={9mm 6mm 12mm 1mm},clip,height=4.1cm]{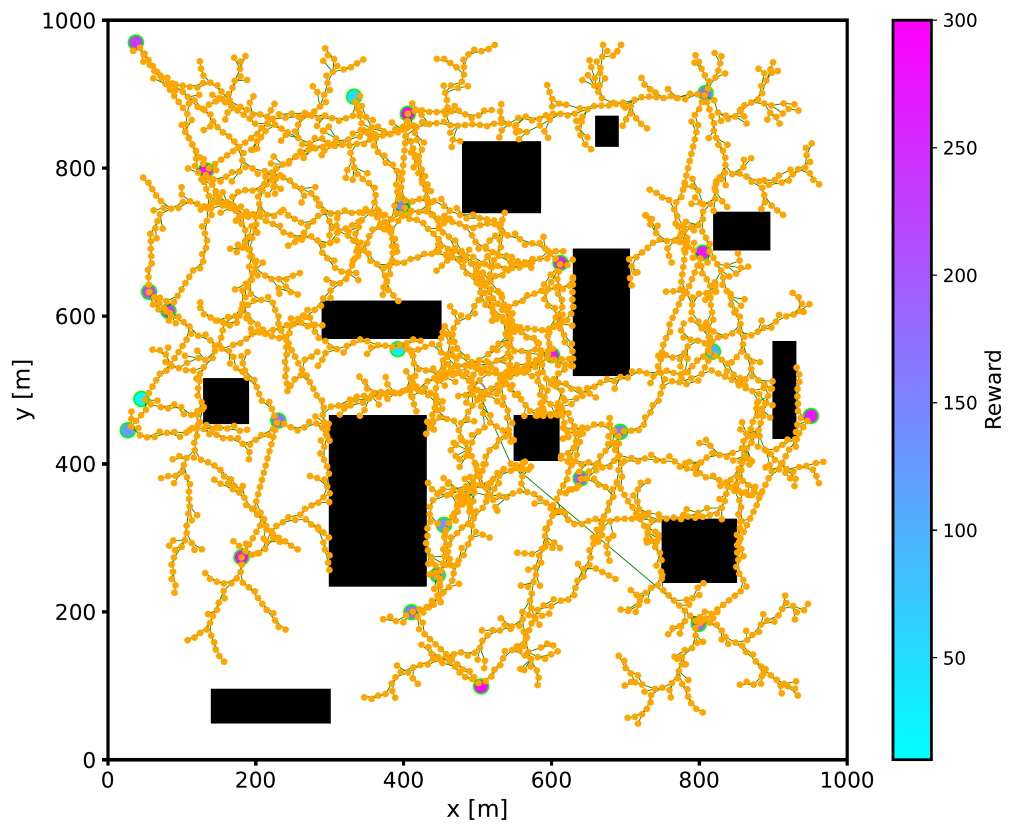}
         \includegraphics[trim={2cm 2cm 2cm 2cm},clip,height=4.1cm]{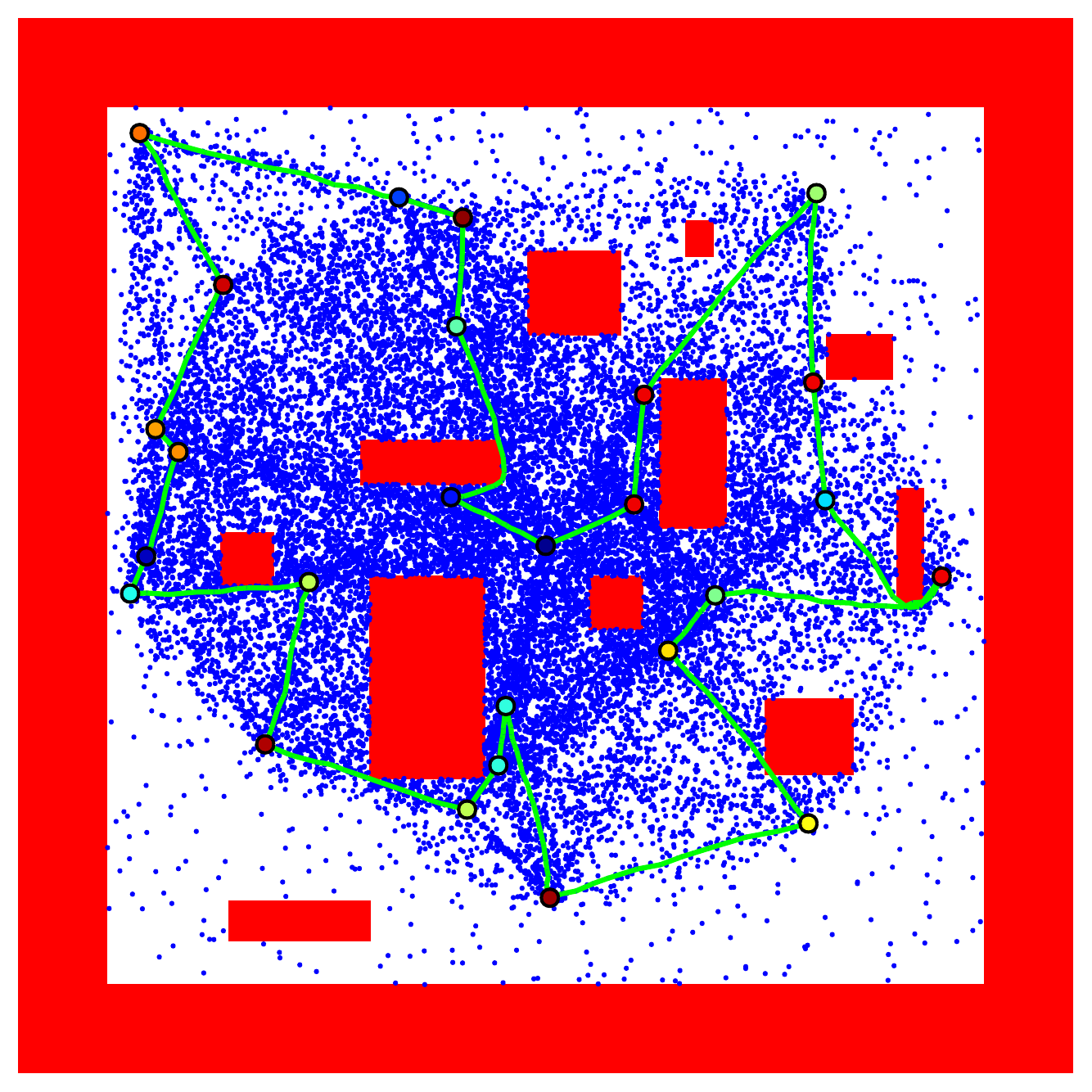}
         \caption{MEDIUM}
     \end{subfigure}
     \begin{subfigure}[b]{\linewidth}
         \centering
         \includegraphics[trim={9mm 6mm 12mm 1mm},clip,height=4.1cm]{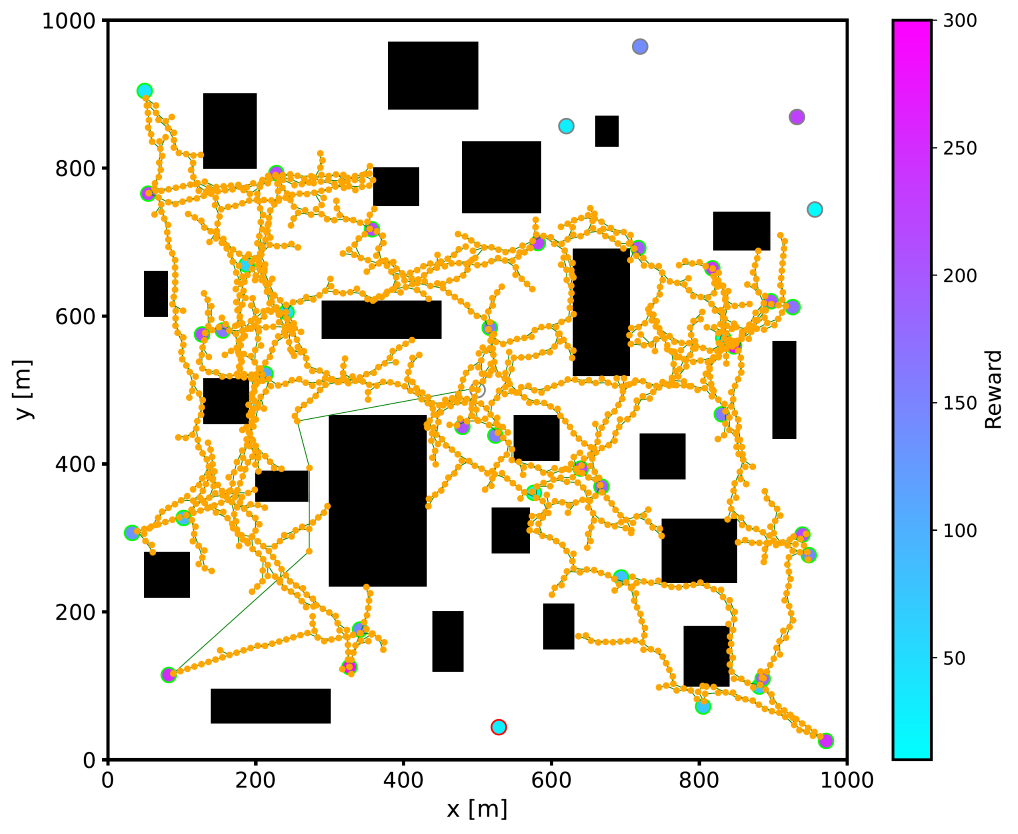}
         \includegraphics[trim={2cm 2cm 2cm 2cm},clip,height=4.1cm]{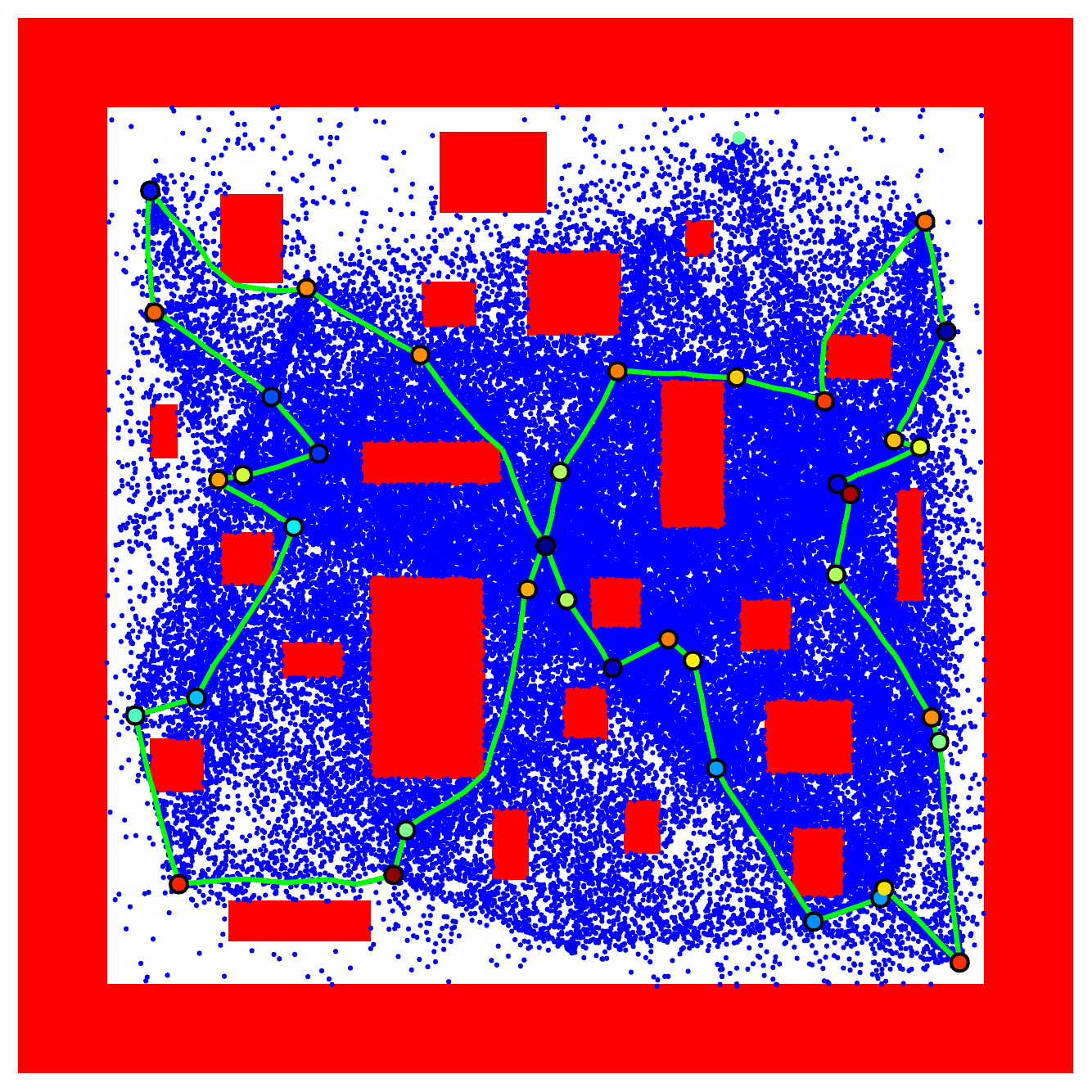}
         \caption{HIGH}
     \end{subfigure}
        \caption{Comparison of the final states obtained by Reward-Guided RRT* (left) and VNS-PRM*~\cite{Penicka2019Physical} (right).}
        \label{fig:comparacao_nodes}
\end{figure}

Our approach achieved the same collected reward as VNS-PRM* in both LOW and MEDIUM scenarios, while collecting only 4\% less in the HIGH. However, a key aspect of this comparison is the sampling density each method requires. The VNS-PRM* solutions are obtained from a substantially denser set of samples, resulting in a considerably larger roadmap than the number of nodes generated by Reward-Guided RRT* during a single run. Figure~\ref{fig:comparacao_nodes} and Table~\ref{tab:comparacao_nodes} provide a visual and quantitative comparison of the number of samples/nodes used by the two approaches. This highlights an important trade-off: VNS-PRM* can achieve slightly better solutions in the most challenging scenario through a much denser state-space representation, whereas Reward-Guided RRT* obtains comparable solutions with substantially fewer generated nodes and without requiring the construction of a dense roadmap.

\begin{table}[htbp]
\centering
\caption{Number of final-tree nodes generated by Reward-Guided RRT* and samples used by VNS-PRM*~\cite{Penicka2019Physical}.}
\label{tab:comparacao_nodes}
\begin{tabular}{c c c}
\toprule
 & Reward-Guided RRT* & VNS-PRM*~\cite{Penicka2019Physical} \\
\midrule
LOW    & 1230 & 13229 \\
MEDIUM & 2925 & 21803 \\
HIGH   & 1695 & 57153 \\
\bottomrule
\end{tabular}
\end{table}

\subsection{Kinematic Constraints}
\label{sec:kinematic}

Finally, to demonstrate the proposed framework's ability to incorporate vehicle kinematic constraints, we extended the planner using Dubins curves as the local steering method \cite{dubins1957curves}. The vehicle state is represented as $\state = (x, y, \theta)$, where $(x, y)$ is the car position and $\theta$ is its orientation. Connections between RRT* nodes are generated subject to a minimum turning radius $R_{\min}$, such that $|\kappa(s)| \leq 1/R_{\min}$. Consequently, edge costs correspond to Dubins-path lengths, and collision checking is performed along the resulting curved paths. The same steering constraint is applied to both the outward and return trees, ensuring that the complete budget-feasible mission is kinematically feasible.

\begin{figure}[htpb]
    \centering
    \begin{subfigure}[b]{.45\linewidth}
         \centering
         \includegraphics[height=3.5cm]{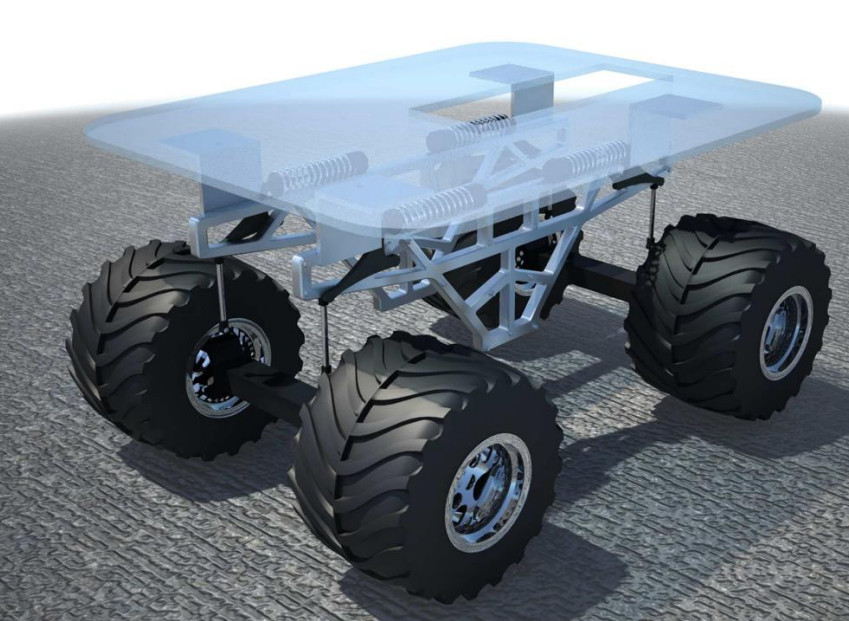}
         \caption{Scaled car-like platform.}
     \end{subfigure}
     ~
     \begin{subfigure}[b]{.45\linewidth}
         \centering
         \includegraphics[height=3.8cm]{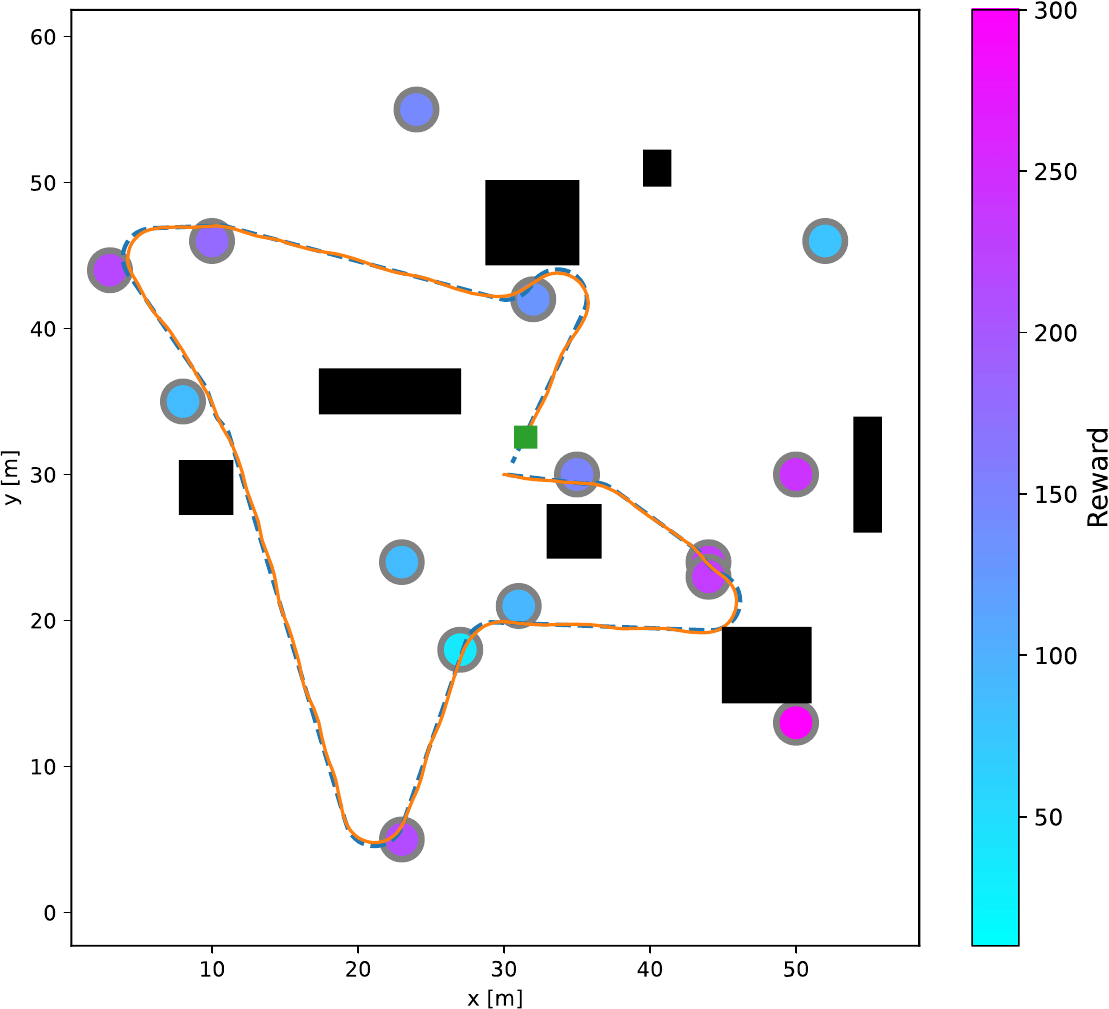}
         \vspace{-6mm}
         \caption{Mission path.}
     \end{subfigure}
      \begin{subfigure}[b]{\linewidth}
         \centering
         \includegraphics[height=3.75cm]{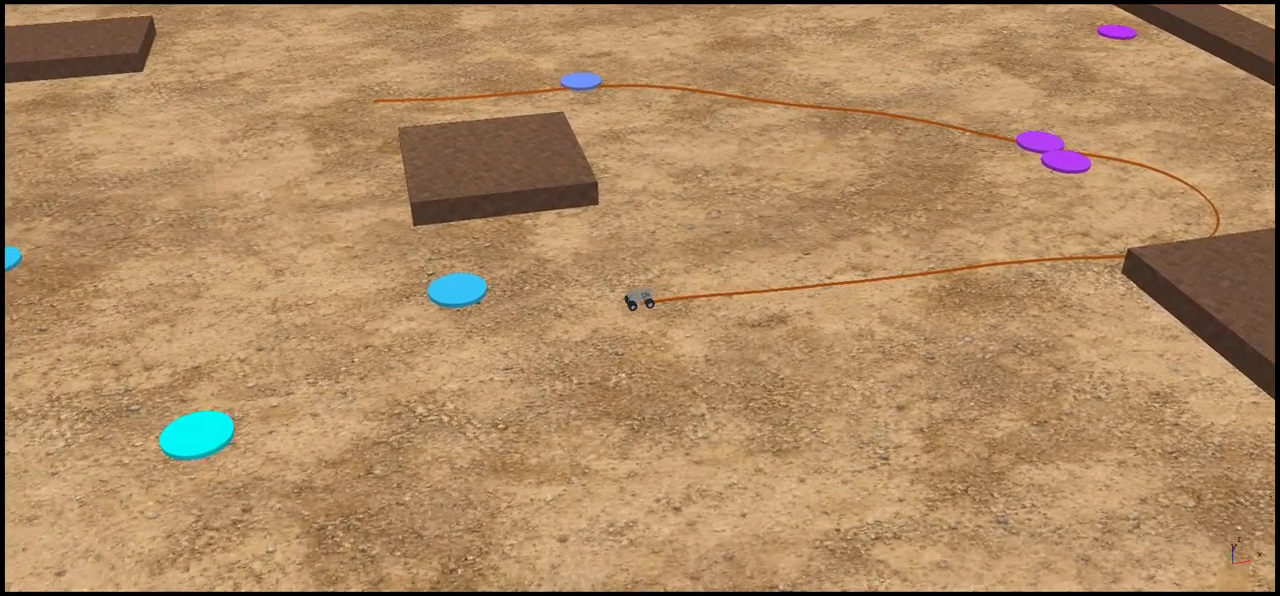}
         \caption{Simulation.}
     \end{subfigure}
    \caption{Kinematic evaluation of the planned mission in CoppeliaSim. (a) Scaled car-like robotic platform used in the experiment; (b) planned route (orange) and path executed (dashed blue) by the vehicle through the selected reward-bearing tasks; (c) simulation environment and execution. Video: \protect\url{https://streamable.com/g7wgkh}}
    \label{fig:dubins_coppelia}
\end{figure}

The resulting path was evaluated in CoppeliaSim\footnote{\url{https://coppeliarobotics.com/}} using the scaled car-like robotic platform shown in Fig.~\ref{fig:dubins_coppelia}(a). The vehicle has a mass of approximately $5\,\mathrm{kg}$ and $R_{\min} \approx 2\,\mathrm{m}$. During the experiment, the vehicle was commanded to travel at $1\,\mathrm{m/s}$ with a travel budget of 250 meters. About 61\% of the total reward was collected in this scenario. Fig.~\ref{fig:dubins_coppelia}(b) shows the planned route and resulting execution of the mission while respecting the vehicle's turning constraint.

We emphasize that this example illustrates the generality of the proposed approach, as planning for a Dubins vehicle in an obstacle-rich environment effectively combines the characteristics of the DOP~\cite{Penicka2017Dubins} and POP~\cite{Penicka2019Physical}.

%%%%%%%%%%%%%%%%%%%%%%%%%%%%%%%%%%%%%%%%%%%%%%%%%%
\section{Conclusion and Future Work}

% The Reward-Guided RRT* proposes a novel approach to solve the Physical Orienteering Problem (POP) by solving the exploration of the state space and task-selection simultaneously. This paper approach uses multiple subsequent RRT* that grows it's branches toward goals based on their reward and distance, and uses path optimization algorithms to reduce budget used while maintaining reward collected and a feasible path. The results highlight the possibility of finding near-optimal solutions without separating path and task-selecting planning in diverse scenarios of varying complexity, with the addition of requiring less samples to achieve such results.

This work introduced the Reward-Guided RRT*, a sampling-based framework that jointly addresses task selection and continuous-space motion planning under a finite budget. The approach incrementally explores the state space while guiding tree expansion toward rewarding locations based on their reward and travel cost, with path optimization reducing the required budget while preserving reward and feasibility.
The experimental results demonstrate the ability to obtain near-optimal solutions across scenarios of varying complexity. Compared with VNS-PRM*, Reward-Guided RRT* achieved the same reward in the LOW and MEDIUM scenarios and only 4\% less in the HIGH scenario while using substantially fewer samples. These results highlight its potential as a general sampling-based framework for budgeted task and motion planning.

% To further improve the Reward-Guided RRT*, the integration of a post final path optimization seems promising, since the path between tasks can be not optimal and budget not spent can be used to test the feasibility to collect new tasks. Also, an implementation of this paper approach in a 3D environment, while considering the budget a more realistic measure instead of distance traveled, such as energy consumed, can be done to prove it's robustness for vehicles that necessitates motion planning in such state space, like UAV's.

Future work includes investigating more advanced mission-level optimization strategies that can reconsider previously committed tasks and exploit the remaining budget more effectively. Extending the framework to 3D environments and replacing traveled distance with more realistic budget measures, such as energy consumption, would also enable evaluation with vehicles such as UAVs, whose motion constraints are naturally represented in higher-dimensional state spaces.
A further direction is to relax the irrevocable-commitment assumption, \eg, through limited backtracking or a rolling-horizon re-evaluation of recently committed segments, to mitigate the greedy nature of the current construction.

%%%%%%%%%%%%%%%%%%%%%%%%%%%%%%%%%%%%%%%%%%%%%%%%%%%%%%%%%%%%%%%%%%%%%%%%%%%%%%%%

%\section*{APPENDIX}

% \section*{Acknowledgment}

%%%%%%%%%%%%%%%%%%%%%%%%%%%%%%%%%%%%%%%%%%%%%%%%%%%%%%%%%%%%%%%%%%%%%%%%%%%%%%%%

\bibliographystyle{IEEEtran}
\bibliography{bibliography}

\end{document}

%% file: defs.tex
% %%%%%%%%%%%%%%%%%%%%%%%%%
% % definicoes
% %%%%%%%%%%%%%%%%%%%%%%%%%
% % operacoes
\newcommand{\escalar}[1]{\ensuremath{\mathit{#1}}}
\newcommand{\vetor}[1]{\ensuremath{\vec{#1}}}
% \newcommand{\matriz}[1]{\ensuremath{\mathbf{\uppercase{#1}}}}
% \newcommand{\conjunto}[1]{\ensuremath{\mathcal{\uppercase{#1}}}}
% \newcommand{\distribuicao}[1]{\ensuremath{\mathcal{\uppercase{#1}}}}
% \newcommand{\transpose}{\ensuremath{{}^\intercal}}
% \newcommand{\espaco}[1]{\ensuremath{\mathds{\MakeUppercase#1}}}
% \newcommand{\complex}[1][\cdot]{\escalar{O}\!\left(#1\right)}

% %%%%%%%%%%%%%%%%%%%%%%%%%
% % vetores

\newcommand{\state}{\vetor{x}}
\newcommand{\stateinit}{\state_{\mathrm{start}}}
\newcommand{\stategoal}{\state_{\mathrm{goal}}}
\newcommand{\stateroot}{\state_{\mathrm{root}}}
\newcommand{\staterand}{\state_{\mathrm{rand}}}
\newcommand{\statenew}{\state_{\mathrm{new}}}
\newcommand{\statenearest}{\state_{\mathrm{nearest}}}